# Defining Relative Likelihood in Partially-Ordered Preferential Structures

**Joseph Y. Halpern**                                        HALPERN@CS.CORNELL.EDU
*Cornell University, Computer Science Department*
*Ithaca, NY 14853*
*http://www.cs.cornell.edu/home/halpern*

## Abstract

Starting with a likelihood or preference order on worlds, we extend it to a likelihood ordering on sets of worlds in a natural way, and examine the resulting logic. Lewis earlier considered such a notion of relative likelihood in the context of studying counterfactuals, but he assumed a total preference order on worlds. Complications arise when examining partial orders that are not present for total orders. There are subtleties involving the exact approach to lifting the order on worlds to an order on sets of worlds. In addition, the axiomatization of the logic of relative likelihood in the case of partial orders gives insight into the connection between relative likelihood and default reasoning.

## 1. Introduction

A *preference order* $\succeq$ on a set $W$ of worlds is a reflexive, transitive relation on $W$. Various readings have been given to the $\succeq$ relation in the literature; $u \succeq v$ has been interpreted as "$u$ at least as preferred or desirable as $v$" (Kraus, Lehmann, & Magidor, 1990; Doyle, Shoham, & Wellman, 1991) (it is this reading that leads to the term "preferential structure"), "$u$ at least as normal (or typical) as $v$" (Boutilier, 1994), and "$u$ is no more remote from actuality than $v$" (Lewis, 1973). In this paper, we focus on one other interpretation, essentially also considered by Lewis (1973). We interpret $u \succeq v$ as meaning "$u$ is at least as likely as $v$".[1] Interestingly, all these readings seem to lead to much the same properties.

In the literature, preference orders have been mainly used to give semantics to conditional logics (Lewis, 1973) and, more recently, to nonmonotonic logic (Kraus et al., 1990). The basic modal operator in these papers has been a conditional $\rightarrow$, where $p \rightarrow q$ is interpreted as "in the most preferred/normal/likely worlds satisfying $p$, $q$ is the case". However, if we view $\succeq$ as representing likelihood, then it seems natural to define a binary operator $\gg$ on formulas such that $\varphi \gg \psi$ is interpreted as "$\varphi$ is more likely than $\psi$". Lewis (1973) in fact did define such an operator, and showed how it related to $\rightarrow$. However, he assumed that $\succeq$ was *total*; that is, he assumed that for all worlds $w, w' \in W$, either $w \succeq w'$ or $w' \succeq w$. But in many cases in preferential or likelihood reasoning, it seems more appropriate to allow the preference order to be partial. It may well be that an agent finds two

---

situations incomparable as far as normality or likelihood goes. For example, one situation may be better in one dimension but worse in another.

As we show in this paper, there are some subtleties involved in starting with a *partial* preference order on worlds. What we are ultimately interested is not an ordering on worlds, but an ordering on *sets* of worlds. To make sense of a statement like $\varphi \gg \psi$, we need to compare the relative likelihood of the set of worlds satisfying $\varphi$ to that of the set satisfying $\psi$. Unfortunately, there are many possible ways of extending a preference order on worlds to one on sets of worlds. We focus here on two particular choices, which both agree with the definition given by Lewis in the case that the preference order on worlds is total, but differ in general. We define $\gg$ using the definition that allows us to make the most interesting to work in default reasoning.

We then turn our attention to axiomatizing the likelihood operator. Lewis provided an axiomatization for the case where the preference order is partial. The key axioms used by Lewis were transitivity:

$$(\varphi_1 \gg \varphi_2) \wedge (\varphi_2 \gg \varphi_3) \Rightarrow (\varphi_1 \gg \varphi_3),$$

and the *union property*:

$$(\varphi_1 \gg \varphi_2) \wedge (\varphi_1 \gg \varphi_3) \Rightarrow (\varphi_1 \gg (\varphi_2 \vee \varphi_3)).$$

This latter property is characteristic of possibility logic (Dubois & Prade, 1990). In the partially ordered case, these axioms do not suffice. We need the following axiom:

$$((\varphi_1 \vee \varphi_2) \gg \varphi_3) \wedge ((\varphi_1 \vee \varphi_3) \gg \varphi_2) \Rightarrow (\varphi_1 \gg (\varphi_2 \vee \varphi_3)).$$

It is not hard to show that this axiom implies transitivity and the union property (in the presence of the other axioms), but it is not equivalent to them. Interestingly, it is the property captured by this axiom that was isolated in (Friedman & Halpern, 1997) as being the key feature needed for a likelihood ordering on sets to be appropriate for doing default reasoning in the spirit of (Kraus et al., 1990). Thus, by considering preference orders that are partial, we are able to clarify the connections between $\gg$, $\rightarrow$, and default reasoning.

The rest of this paper is organized as follows. In Section 2, we consider how to go from an ordering on worlds to one on sets of worlds, focusing on the differences between total and partial preference orders. In Section 3, we present a logic for reasoning about relative likelihood and provide a natural complete axiomatization for it. In Section 4, we relate our results to other work on relative likelihood, as well as to work on conditional logic and nonmonotonic reasoning. We conclude in Section 5. Proofs of the technical results can be found in Appendix A.

## 2. From Preorders on Worlds to Preorders on Sets of Worlds

We capture the likelihood ordering on a set $W$ of possible worlds by a *partial preorder*—that is, a reflexive and transitive relation—$\succeq$ on $W$.[2] We typically write $w' \succeq w$ rather than $(w', w) \in \succeq$. As usual, we often write $u \preceq v$ rather than $v \succeq u$, and take $u \succ v$ to be an abbreviation for $u \succeq v$ and not($v \succeq u$), and $u \prec v$ to be an abbreviation for $v \succ u$. The relation $\succ$ is a *strict partial order*, that is, it is an irreflexive (for all $w$, it is not the case that $w \succ w$) and transitive relation on $W$. We say that $\succ$ is the strict partial order *determined by* $\succeq$.

---

2. A partial order $R$ is typically assumed to be reflexive, transitive, and *anti-symmetric* (so that if $(a, b) \in R$ and $(b, a) \in R$, then $a = b$). We are not assuming that $\succeq$ is anti-symmetric here, which is why it is a *preorder*.

As we said in the introduction, we think of $\succeq$ as providing a likelihood, or preferential, order on the worlds in $W$. Thus, $w \succeq w'$ holds if $w$ is at least as likely/preferred/normal/close to actuality as $w'$. Given this interpretation, the fact that $\succeq$ is assumed to be a partial preorder is easy to justify. For example, transitivity just says that if $u$ is at least as likely as $v$, and $v$ is at least as likely as $w$, then $u$ at least as likely as $w$. Notice that since $\succeq$ is a *partial* preorder, there may be some pairs of worlds $w$ and $w'$ that are incomparable according to $\succeq$. Intuitively, we may not be prepared to say that either one is likelier than the other. We say that $\succeq$ is a *total preorder* (or *connected*, or a *linear preorder*) if for all worlds $w$ and $w'$, either $w \succeq w'$ or $w' \succeq w$.

Since we have added likelihood to the worlds, it seems reasonable to also add likelihood to the language, to allow us to say "$\varphi$ is more likely than $\psi$", for example. But what exactly should this mean? Although having $\succeq$ in our semantic model allows us to say that one world is more likely than another, it does not immediately tell us how to say that a set of worlds is more likely than another set. But, as we observed in the introduction, this is just what we need to make sense of "$\varphi$ is more likely than $\psi$".

How should we extend a likelihood ordering on worlds to one on sets of worlds? Clearly we want to do so in a way that preserves the ordering on worlds. That is, using $\rhd$ to denote the ordering sets, we would certainly expect that $u \succeq v$ would imply $\{u\} \rhd \{v\}$. We could impose a few other minimal requirements, but they certainly would not be enough to uniquely determine an ordering on sets. For example, here are two general approaches; to distinguish them, we put subscripts on $\rhd$.

1. Define $\rhd_1$ so that $U \rhd_1 V$ if for all $u \in U$ and all $v \in V$, we have $u \succeq v$. We can define a strict partial order in this spirit in two distinct ways.

    (a) By considering the analogous procedure to that used to get $\succ$ from $\succeq$: We define $\rhd_2$ so that $U \rhd_2 V$ if $U \rhd_1 V$ and not($V \rhd_1 U$). We call this the *standard method* below.

    (b) By replacing $\succeq$ in the definition of $\rhd_1$ by $\succ$: We define $\rhd_3$ so that $U \rhd_3 V$ if for all $u \in U$ and all $v \in V$, we have $u \succ v$. We call this the *alternative method*.

2. Define $\rhd_4$ so that $U \rhd_4 V$ if for all $v \in V - U$, there is some $u \in U - V$ such that $u \succeq v$. Note that this approach focuses on the symmetric difference between $U$ and $V$. There are again two ways of getting a strict partial order.

    (a) The standard method gives us $U \rhd_5 V$ if $U \rhd_4 V$ and not($V \rhd_4 U$).

    (b) The alternative method gives us $U \rhd_6 V$ if for all $v \in V - U$ there is some $u \in U - V$ such that $u \succ v$.

The first approach ($\rhd_1$) was used by Doyle, Shoham, and Wellman (1991) in defining a logic of relative desire, starting with a preference order on worlds. Unfortunately, as they themselves point out, these relations are too weak to allow us to make important distinctions. They go on to define other notions of comparison, but these are more tuned to their applications, and not in the spirit of the notions we are considering here.

The second approach (typically $\triangleright_4$ and $\triangleright_5$) has been widely used in various applications in the literature. For example,

- Dershowitz and Manna (1979) use it to define on ordering on multisets, which is then used to provide a technique for proving program termination.

- Przymusinski (1987) uses it to order models of a database.

- Brass (1991), Cayrol, Royer, and Saurel (1992), Delgrande (1994), and Geffner (1992) all use it to help model various aspects of default reasoning.

In this paper, we focus on a variant of the second approach, essentially due to Lewis (1973), which has interesting connections to default reasoning. Roughly speaking, we take $U$ to be more likely than $V$ if for every world in $V$, there is a more likely world in $U$.

To make this precise, if $U, V \subseteq W$, we write $U \succeq^s V$ if for every world $v \in V$, there is a world $u \in U$ such that $u \succeq v$. It is easy to check that $\succeq^s$ is a partial preorder, that is, it is reflexive and transitive. (The superscript $s$ is for "set".) Moreover, if $\succeq$ is a total preorder, then so is $\succeq^s$. Finally, as we would expect, we have $u \succeq v$ iff $\{u\} \succeq^s \{v\}$, so the $\succeq^s$ relation on sets of worlds can be viewed as a generalization of the $\succeq$ relation on worlds.

We can then apply both the standard method and the alternative method to define a strict partial order. The standard method gives us the relation $\succ'$, where $U \succ' V$ holds if $U \succeq^s V$ and not($V \succeq^s U$). The alternative method gives us the relation $\succ^s$ defined on finite sets by taking $U \succ^s V$ to hold if $U$ is nonempty, and for every world $v \in V$, there is a world $u \in U$ such that $u \succ v$. (The reasons that $U$ is taken to be nonempty and that the definition is restricted to finite sets are discussed below.)

How do these various approaches compare? Clearly $U \triangleright_1 V$ implies $U \succeq^s V$, although the converse does not hold in general; $\triangleright_1$ is a very weak ordering. As a consequence, $\triangleright_2$ and $\succ'$ are incomparable, as are $\triangleright_3$ and $\succ^s$. Similar remarks apply to $\triangleright_4$. Again, it is easy to see that $U \triangleright_4 V$ implies $U \succeq^s V$. On the other hand, the two notions are not equivalent. For example, suppose $v \succ v'$. Then $\{v\} \succeq^s \{v, v'\}$, but we do not have $\{v\} \triangleright_4 \{v, v'\}$.

Why are we focusing on $\succeq^s$, $\succ'$, and $\succ^s$ here, rather than $\triangleright_1-\triangleright_6$ (or some other notion)? From a likelihood viewpoint, they all seem to be reasonable; our intuitions regarding extending likelihood from worlds to sets of worlds do not seem to be that well developed. It may be possible to motivate $\succeq^s$ as the finest relation extending $\succeq$ that has certain properties, but that was not in fact our motivation here. Rather, our interest is motivated by the deep connections between $\succ'$ and $\succ^s$ and certain approaches to nonmonotonic reasoning. Having said that, many of the questions that we consider here could perfectly well be explored using $\triangleright_1-\triangleright_6$. With that apology, we do not discuss $\triangleright_1-\triangleright_6$ further in this paper, except for the odd remark.

We have explained that there are two methods for getting a strict partial order on sets of worlds from a partial order on worlds. But are the standard method and the alternative method really so different? It is easy to see that $u \succ v$ iff $\{u\} \succ^s \{v\}$ iff $\{u\} \succ' \{v\}$. (This is true for $\triangleright_2$, $\triangleright_3$, $\triangleright_5$, and $\triangleright_6$ as well.) Thus, $\succ^s$ and $\succ'$ agree on singleton sets and extend the $\succ$ relation on worlds. Moreover, both $\succ^s$ and $\succ'$ are strict partial orders on finite sets. (The requirement that $U$ must be nonempty in the definition of $U \succ^s V$ is there to ensure that we do not have $\emptyset \succ^s \emptyset$; strictly speaking, we should have also added it to the definitions of $\rhd_3$ and $\rhd_6$ to ensure that they were strict partial orders.) As shown in Lemma 2.9, $\succ'$ and $\succ^s$ are in fact identical if the underlying preorder $\succeq$ on worlds is a total preorder. However, as the following example shows, $\succ^s$ and $\succ'$ are not identical in general.

**Example 2.1:** Suppose $W = \{w_1, w_2\}$, and $\succeq$ is such that $w_1$ and $w_2$ are incomparable. Then it is easy to see that $\{w_1, w_2\} \succ' \{w_1\}$. However, it is not the case that $\{w_1, w_2\} \succ^s \{w_1\}$, since there is no element of $\{w_1, w_2\}$ that is strictly more likely than $w_1$.[3] ▨

Notice that we were careful to define $\succ$ as we did only on finite sets. The following example illustrates why:

**Example 2.2:** Let $W_\infty = \{w_0, w_1, w_2, \ldots\}$, and suppose that $\succeq$ is such that

$$w_0 \prec w_1 \prec w_2 \prec \ldots$$

Then it is easy to see that if we were to apply the definition of $\succ^s$ to infinite sets, then we would have $W_\infty \succ^s W_\infty$, and $\succ^s$ would not be irreflexive.[4] ▨

The approach for extending the definition of $\succ^s$ to infinite sets is essentially due to Lewis (1973) (who did it for the case of total preorders). The idea is to say that in order to have $U \succ^s V$, it is not enough that for every element $v$ in $V$ there is some element $u$ in $U$ that is more likely than $v$. This definition is what allows $W_\infty \succ W_\infty$ in Example 2.2. Notice that in the finite case, it is easy to see that if $U \succ^s V$, then for every element $v$ in $V$, there must some $u \in U$ such that, not only do we have $u \succ v$, but $u$ *dominates* $V$ in that for no $v' \in V$ do we have $v' \succ u$. It is precisely this domination condition that does not hold in Example 2.2. This observation provides the motivation for the official definition of $\succ^s$, which applies in both finite and infinite domains.

**Definition 2.3:** Suppose $\succeq$ is a partial preorder on $W$, $U, V \subseteq W$, and $w \in W$. We say that $w$ *dominates* $V$ if for no $v \in V$ is it the case that $v \succ w$. (Notice that if $\succeq$ is a total preorder, this is equivalent to saying that $w \succeq v$ for all $v \in V$.) We write $U \succ^s V$ if $U$ is nonempty and, for all $v \in V$, there exists $u \in U$ such that $u \succ v$ and $u$ dominates $V$. ▨

It is easy to see that this definition of $\succ^s$ agrees with our earlier definitions if $U$ and $V$ are finite.

We now collect some properties of $\succ^s$, $\succ'$, and $\succeq^s$. To do this, we need a few definitions. We say that a relation $\rhd$ on $2^W$ (not necessarily a preorder) is *qualitative* if $(V_1 \cup V_2) \rhd V_3$ and $(V_1 \cup V_3) \rhd V_2$ implies $V_1 \rhd (V_2 \cup V_3)$. We say that $\rhd$ *satisfies the union property* if $V_1 \rhd V_2$ and $V_1 \rhd V_3$ implies $V_1 \rhd (V_2 \cup V_3)$. We say that $\rhd$ is *orderly* if $U \rhd V$, $U' \supseteq U$, and $V' \subseteq V$ implies $U' \rhd V'$. We provide some intuition for these properties following Proposition 2.5, after showing how they help us characterize $\succ^s$, $\succ'$, and $\succeq^s$.

---

3. We remark that similar results hold for $\rhd_5$ and $\rhd_6$. They are identical if the underlying order on worlds is a total preorder, and this example can also be used to show that they differ if the underlying order is partial.

4. Similar problems arise for $\rhd_6$ when dealing with infinite sets, and the solution for $\succ^s$ described in Definition 2.3 can be applied to $\rhd_6$ as well.

**Lemma 2.4:** *If $\rhd$ is an orderly qualitative relation on $2^W$, then $\rhd$ is transitive and satisfies the union property.*

**Proof:** See Appendix A. ■

The converse to Lemma 2.4 does not hold. Indeed, an orderly strict partial order on $2^W$ may satisfy the union property and still not be qualitative. For example, suppose $W = \{a, b, c\}$, and we have $\{a, b\} \rhd \{c\}$, $\{a, c\} \rhd \{b\}$, $\{a, b, c\} \rhd \{b\}$, $\{a, b, c\} \rhd \{c\}$, and $\{a, b, c\} \rhd \{b, c\}$. It can easily be checked that $\rhd$ is an orderly strict partial order that satisfies the union property, but is not qualitative.

With these definitions in hand, we can state the key properties of the relations we are interested in here.

**Proposition 2.5:**

(a) If $\succeq$ is a partial preorder on $W$, then $\succeq^s$ is an orderly partial preorder on $2^W$ that satisfies the union property.

(b) If $\succeq$ is a partial preorder on $W$, then $\succ'$ is an orderly strict partial order on $2^W$.

(c) If $\succ$ is a strict partial order on $W$, then $\succ^s$ is an orderly qualitative strict partial order on $2^W$.

**Proof:** See Appendix A. ■

We can now discuss how to interpret the properties we have been considering in light of this result.

To the extent that we think of $\rhd$ as meaning "more likely than", then orderliness is a natural property to require. If $U$ is more likely than $V$, then certainly any superset of $U$ should be more likely than any subset of $V$. It is thus not surprising that all three of the relations we have defined are orderly.

Clearly the union property generalizes to arbitrary finite unions. That is, if $\rhd$ satisfies the union property and $A \rhd B_i$, $i = 1, \ldots, n$, then $A \rhd B_1 \cup \ldots \cup B_n$. In particular, if $u \succeq v_j$ for $j = 1, \ldots, N$, then $\{u\} \succeq^s \{v_1, \ldots, v_N\}$, no matter how large $N$ is, and similarly if we replace $\succeq$ by $\succ$ (since the fact that $\succ^s$ is qualitative means that it satisfies the union property, by Lemma 2.4). This is very different from probability, where sufficiently many "small" probabilities eventually can dominate a "large" probability. This suggests that $u \succ v$ should perhaps be interpreted as "$u$ is *much* more likely than $v$". More generally, if $\rhd$ satisfies the union property, then $U \rhd V$ can be interpreted as meaning that $U$ is much more likely than $V$. In this sense, the notion of likelihood corresponding to $\succ^s$ or $\succeq^s$ is closer to possibility (Dubois & Prade, 1990) than probability, since the relation "more possible than" satisfies the union property.

Note that, in general, $\succ'$ does *not* satisfy the union property. In Example 2.1, we have $\{w_1, w_2\} \succ' \{w_1\}$ and $\{w_1, w_2\} \succ' \{w_2\}$, but we do not have $\{w_1, w_2\} \succ' \{w_1, w_2\}$.

The qualitative property is somewhat more difficult to explain intuitively. Of the three relations we are considering, only the $\succ^s$ relation satisfies it. The fact that $\succ'$ does not satisfy it follows from Lemma 2.4, together with the observation that $\succ'$ does not satisfy the union property. Example 2.1 also shows that $\succeq^s$ is not qualitative, since if it were, we could conclude from $\{w_1, w_2\} \succeq^s \{w_2\}$ (taking $V_1 = \{w_1\}$ and $V_2 = V_3 = \{w_2\}$ in the definition of qualitative) that $\{w_1\} \succeq^s \{w_2\}$, a contradiction. Our interest in the qualitative property stems from the fact that, in a precise sense, it is the property that characterizes $\succ^s$. It first arose in (Friedman & Halpern, 1997), where it was shown to be the key property required of a generalization of probability called *plausibility* to capture default reasoning. This is discussed in more detail in Section 4.

If $\succeq$ is a total preorder, then we get further connections between these notions. Before we discuss the details, we need to define the analogue of total preorders in the strict case. A relation $\rhd$ on an arbitrary set $W'$ (not necessarily of the form $2^W$) is *modular* if $w_1 \rhd w_2$ implies that, for all $w_3$, either $w_3 \rhd w_2$ or $w_1 \rhd w_3$. Modularity is the "footprint" of a total preorder on the strict order determined by it. This is made precise in the following lemma.

**Lemma 2.6:** *If $\succeq$ is a total preorder, then the strict partial order $\succ$ determined by $\succeq$ is modular. Moreover, if $\rhd$ is a modular, strict partial order on $W$, then there is a total preorder $\succeq$ on $W$ such that $\rhd$ is the strict partial order determined by $\succeq$.*

**Proof:** See Appendix A. ∎

Modularity is preserved when we lift the preorder from $W$ to $2^W$.

**Lemma 2.7:** *If $\succ$ is a modular relation on $W$, then $\succ^s$ is a modular relation on $2^W$.*

**Proof:** See Appendix A. ∎

Although we showed that the converse to Lemma 2.4 does not hold in general for strict partial orders, it does hold for orders that are modular.

**Lemma 2.8:** *If $\rhd$ is a modular strict partial order and satisfies the union property, then $\rhd$ is qualitative.*

**Proof:** See Appendix A. ∎

As shown in (Friedman & Halpern, 1997), there is a connection between nonmonotonic reasoning, conditional logic, and the qualitative property. (This is discussed in Section 4.) This relationship is best understood by considering $\succ^s$, rather than $\succeq^s$ or $\succ'$, which is why we focus on $\succ^s$ here. Lewis (1973) was able to use $\succ'$ because he focused on total preorders. The following lemma makes this precise.

**Lemma 2.9:** *If $\succeq$ is a total preorder, then $\succ^s$ and $\succ'$ agree. In general, $U \succ^s V$ implies $U \succ' V$, but the converse does not hold.*

**Proof:** See Appendix A. ∎

We close this section by considering when a preorder on $2^W$ can be viewed as being generated by a preorder on $W$. This result turns out to play a key role in our completeness proof, and emphasizes the role of the qualitative property.

**Theorem 2.10:** *Let $\mathcal{F}$ be a finite algebra of subsets of $W$ (that is, $\mathcal{F}$ is a set of subsets of $W$ that is closed under union and complementation and contains $W$ itself) and let $\rhd$ be an orderly qualitative relation on $\mathcal{F}$.*

(a) If $\triangleright$ is a total preorder on $\mathcal{F}$, then there is a total preorder $\succeq$ on $W$ such that $\triangleright$ and $\succeq^s$ agree on $\mathcal{F}$ (that is, for $U, V \in \mathcal{F}$, we have $U \triangleright V$ iff $U \succeq^s V$).

(b) If $\triangleright$ is a strict partial order and each nonempty set in $\mathcal{F}$ has at least $2^{\log(|\mathcal{F}|)^{\log(|\mathcal{F}|)}}$ elements, then there is a partial preorder $\succeq$ on $W$ such that $\triangleright$ and $\succ^s$ agree on $\mathcal{F}$.

**Proof:** An *atom* of $\mathcal{F}$ is a minimal nonempty element of $\mathcal{F}$. Since $\mathcal{F}$ is finite, it is easy to see that every element of $\mathcal{F}$ can be written as a union of atoms, and the atoms are disjoint. Part (a) is easy: for each $w \in W$, let $A_w$ be the unique atom in $\mathcal{F}$ containing $w$. Define $\succeq$ on $W$ so that $v \succeq w$ iff $A_v \triangleright A_w$. It is easy to see that if $\triangleright$ is a total preorder on $\mathcal{F}$, then $\succeq$ is a total preorder on $W$ and $\triangleright$ agrees with $\succeq^s$ on $\mathcal{F}$. The proof of (b) is considerably more difficult; see Appendix A for details. ■

It is not clear that the requirement that the sets in $\mathcal{F}$ have at least $2^{\log(|\mathcal{F}|)^{\log(|\mathcal{F}|)}}$ elements is necessary. However, it can be shown that Theorem 2.10(b) does not hold without some assumptions on the cardinality of elements in $\mathcal{F}$. For example, suppose that the atoms of $\mathcal{F}$ are $A$, $B$, and $C$. Let $\triangleright$ be defined so that $(B \cup C) \triangleright A$, $W \triangleright A$, $X \triangleright \emptyset$ for all nonempty $X \in \mathcal{F}$, and these are the only pairs of sets that are in the $\triangleright$ relation. It is easy to see that $\triangleright$ is an orderly, qualitative, strict partial order. However, if $W = \{a, b, c\}$, $A = \{a\}$, $B = \{b\}$, and $C = \{c\}$, there is no ordering $\succ$ on $W$ such that $\succ^s$ and $\triangleright$ agree on $\mathcal{F}$: it is easy to see that such an ordering $\succ$ must make $a$, $b$, and $c$ incomparable. But if they are incomparable, we cannot have $\{b, c\} \succ^s \{a\}$. On the other hand, if we allow $C$ to have two elements, by taking $W = \{a, b, c, d\}$, $A = \{a\}$, $B = \{b\}$, and $C = \{c, d\}$, then there is an ordering $\succ$ such that $\succ^s = \triangleright$: we simply take $a \succ c$ and $b \succ d$.

## 3. A Logic of Relative Likelihood

We now consider a logic for reasoning about relative likelihood. Let $\Phi$ be a set of primitive propositions. A *basic likelihood formula (over $\Phi$)* is one of the form $\varphi \gg \psi$, where $\varphi$ and $\psi$ are propositional formulas over $\Phi$. We read $\varphi \gg \psi$ as "$\varphi$ is more likely than $\psi$". Let $\mathcal{L}$ consist of Boolean combinations of basic likelihood formulas. Notice that we do not allow nesting of likelihood in $\mathcal{L}$, nor do we allow purely propositional formulas. There would no difficulty extending the syntax and semantics to deal with them, but this would just obscure the issues of interest here.

A *preferential structure* (over $\Phi$) is a tuple $M = (W, \succeq, \pi)$, where $W$ is a (possibly infinite) set of possible worlds, $\succeq$ is a partial preorder on $W$, and $\pi$ associates with each world in $W$ a truth assignment to the primitive propositions in $\Phi$. Notice that there may be two or more worlds with the same truth assignment. As we shall see, in general, we need to have this, although in the case of total preorders, we assume without loss of generality that there is at most one world associated with each truth assignment.

We can give semantics to formulas in $\mathcal{L}$ in preferential structures in a straightforward way. For a propositional formula $\varphi$, let $[\![\varphi]\!]_M$ consist of the worlds in $M$ whose truth assignment satisfies $\varphi$. We then define

$$M \models \varphi \gg \psi \text{ if } [\![\varphi]\!]_M \succ^s [\![\psi]\!]_M.$$

We extend $\models$ to Boolean combinations of basic formulas in the obvious way.

Notice that $M \models \neg(\neg\varphi \gg false)$ iff $[\![\neg\varphi]\!]_M = \emptyset$ iff $[\![\varphi]\!]_M = W$. Let $K\varphi$ be an abbreviation for $\neg(\neg\varphi \gg false)$. It follows that $M \models K\varphi$ iff $\varphi$ is true at all possible worlds.[5]

With these definitions, we can provide a sound and complete axiomatization for this logic of relative likelihood. Let AX consist of the following axioms and inference rules.

L1. All substitution instances of tautologies of propositional calculus

L2. $\neg(\varphi \gg \varphi)$

L3. $((\varphi_1 \vee \varphi_2) \gg \varphi_3) \wedge ((\varphi_1 \vee \varphi_3) \gg \varphi_2) \Rightarrow (\varphi_1 \gg (\varphi_2 \vee \varphi_3))$

L4. $(K(\varphi \Rightarrow \varphi') \wedge K(\psi' \Rightarrow \psi) \wedge (\varphi \gg \psi)) \Rightarrow \varphi' \gg \psi'$

Gen. $K\varphi$, for all propositional tautologies $\varphi$ (Generalization)

MP. From $\varphi$ and $\varphi \Rightarrow \psi$ infer $\psi$ (Modus ponens)

L2, L3, and L4 just express the fact that $\succ^s$ is irreflexive, qualitative, and orderly, respectively; this is made precise in the proof of the following result. The axiom Gen is the analogue of the inference rule "From $\varphi$ infer $K\varphi$", typically known as generalization. We do not have this inference rule here, since our language does not allow nested occurrences of $\gg$. Thus, for an arbitrary formula $\varphi$, the formula $K\varphi$ is not in our language. It is in our language only if $\varphi$ is propositional; the axiom takes care of this case.

**Theorem 3.1:** *AX is a sound and complete axiomatization of the language $\mathcal{L}$ with respect to preferential structures.*

**Proof:** The validity of L1 is immediate. It is clear that the fact that $\succ^s$ is irreflexive and qualitative, as shown in Proposition 2.5, implies that L2 and L3 are valid. To see that L4 corresponds to orderliness, note that if $M \models K(\varphi \Rightarrow \varphi') \wedge K(\psi' \Rightarrow \psi)$ and $\varphi \gg \psi$, then $[\![\varphi]\!]_M \subseteq [\![\varphi']\!]_M$, $[\![\psi']\!]_M \subseteq [\![\psi]\!]_M$, and $[\![\varphi]\!]_M \succ^s [\![\psi]\!]_M$. Since $\succ^s$ is orderly, it follows that $[\![\varphi']\!]_M \succ^s [\![\psi']\!]_M$, so $M \models \varphi' \gg \psi'$. Thus, L4 is valid. It is also clear that MP and Gen preserve validity. Thus, the axiomatization is sound.

The completeness proof starts out, as is standard for completeness proofs in modal logic, with the observation that it suffices to show that a consistent formula is satisfiable. That is, we must show that every formula $\varphi$ for which it is not the case that $\neg\varphi$ is provable from AX is satisfiable in some preferential structure $M$. However, the standard modal logic techniques of constructing a *canonical* model (see, for example, (Hughes & Cresswell, 1968)) do not seem to work in this case. Finding an appropriate partial preorder on worlds is nontrivial. For this we use (part (b) of) Theorem 2.10. See Appendix A for the details. ∎

What happens if we start with a total preorder? Let $\text{AX}^M$ consist of AX together with the obvious axiom expressing modularity:

L5. $(\varphi_1 \gg \varphi_2) \Rightarrow ((\varphi_1 \gg \varphi_3) \vee (\varphi_3 \gg \varphi_2))$

We say that a preferential structure is *totally preordered* if it has the form $(W, \succeq, \pi)$, where $\succeq$ is a total preorder on $W$.

---

5. $K$ was defined by Lewis (1973), although he wrote $\Box$ rather than $K$.

**Theorem 3.2:** *$AX^M$ is a sound and complete axiomatization of the language $\mathcal{L}$ with respect to totally preordered preferential structures.*

**Proof:** For soundness, we just have to check that L5 is valid in totally ordered preferential structures. This is straightforward and left to the reader. The completeness proof uses Theorem 2.10 again, but is simpler than the proof of completeness in Theorem 3.1. We leave details to Appendix A. ∎

We remark that in light of Proposition 2.8, we can replace L4 in $AX^M$ by axioms saying that $\gg$ is transitive and satisfies the union property, namely:

L6. $(\varphi_1 \gg \varphi_2) \wedge (\varphi_2 \gg \varphi_3) \Rightarrow (\varphi_1 \gg \varphi_3)$

L7. $((\varphi_1 \gg \varphi_2) \vee (\varphi_1 \gg \varphi_3)) \Rightarrow (\varphi_1 \gg (\varphi_1 \vee \varphi_2))$

The result is an axiomatization that is very similar to that given by Lewis (1973).

In the proof of Theorem 3.1, when showing that a consistent formula $\varphi$ is satisfiable, the structure constructed may have more than one world with the same truth assignment. This is necessary, as the following example shows. (We remark that this observation is closely related to the cardinality requirements in Theorem 2.10(b).)

**Example 3.3:** Suppose $\Phi = \{p, q\}$. Let $\varphi$ be the formula $(p \gg (\neg p \wedge q)) \wedge \neg((p \wedge q) \gg (\neg p \wedge q)) \wedge \neg((p \wedge \neg q) \gg (\neg p \wedge q))$. It is easy to see that $\varphi$ is satisfied in a structure consisting of four worlds, $w_1, w_2, w_3, w_4$, such that $w_1 \succ w_3$, $w_2 \succ w_4$, $p \wedge q$ is true at $w_1$, $p \wedge \neg q$ is true at $w_2$, and $\neg p \wedge q$ is true at both $w_3$ and $w_4$. However, $\varphi$ is not satisfiable in any structure where there is at most one world satisfying $\neg p \wedge q$. For suppose $M$ were such a structure, and let $w$ be the world in $M$ satisfying $\neg p \wedge q$. Since $M \models p \gg (\neg p \wedge q)$, it must be the case that $[\![p]\!]_M \succ^s \{w\}$. Thus, there must be a world $w' \in [\![p]\!]_M$ such that $w' \succ w$. But $w'$ must satisfy one of $p \wedge q$ or $p \wedge \neg q$, so $M \models ((p \wedge q) \gg (\neg p \wedge q)) \vee ((p \wedge \neg q) \gg (\neg p \wedge q))$, contradicting the assumption that $M \models \varphi$. ∎

It is not hard to see that the formula $\varphi$ of Example 3.3 is not satisfiable in a totally preordered preferential structure. This is not an accident.

**Proposition 3.4:** *If a formula is satisfiable in a totally preordered preferential structure, then it is satisfiable in a totally preordered preferential structure with at most one world per truth assignment.*

**Proof:** See Appendix A. ∎

The results of this and the previous section help emphasize the differences between totally preordered and partially preordered structures.

## 4. Related Work

The related literature basically divides into two groups (with connections between them): (a) other approaches to relative likelihood and (b) work on conditional and nonmonotonic logic.

We first consider relative likelihood. Gärdenfors (1975) considered a logic of relative likelihood, but he took as primitive a total preorder on the sets in $2^W$, and focused on connections with probability. In particular, he added axioms to ensure that, given a preorder $\succeq^s$ on $2^W$, there was a probability function Pr with the property that (in our notation) $U \succeq^s V$ iff $\Pr(U) \geq \Pr(V)$. Fine (1973) defines a qualitative notion $\preceq$ of *comparative probability*, but like Gärdenfors, assumes that the preorder on sets is primitive, and is largely concerned with connections to probability.

Halpern and Rabin (1987) consider a logic of likelihood where absolute statements about likelihood can be made ($\varphi$ is likely, $\psi$ is somewhat likely, and so on), but there is no notion of relative likelihood.

Of course, there are many more quantitative notions of likelihood, such as probability, possibility (Dubois & Prade, 1990), ordinal conditional functions (OCFs) (Spohn, 1988), and Dempster-Shafer belief functions (Shafer, 1976). The ones closest to the relative likelihood considered here are possibility and OCFs. Recall that a possibility measure Poss on $W$ associates with each world its *possibility*, a number in $[0, 1]$, such that for $V \subseteq W$, we have $\text{Poss}(V) = \sup\{\text{Poss}(v) : v \in V\}$, with the requirement that $\text{Poss}(W) = 1$. Clearly a possibility measure places a total preorder on sets, and satisfies the union property, since $\text{Poss}(A \cup B) = \max(\text{Poss}(A), \text{Poss}(B))$. The same is true for OCFs; we refer the reader to (Spohn, 1988) for details. Fariñas del Cerro and Herzig (1991) define a logic QPL (Qualitative Possibilistic Logic) with a modal operator $\geq$, where $\varphi \geq \psi$ is interpreted as $\text{Poss}([\![\varphi]\!]) \leq \text{Poss}([\![\psi]\!])$. Clearly, $\varphi \geq \psi$ essentially corresponds to $\psi \gg \varphi$. They provide a complete axiomatization for their logic, and prove that it is equivalent to Lewis' logic.[6] Not surprisingly, an analogue of $AX^M$ is also complete for the logic. Further discussion of the logic can be found in (Bendová & Hájek, 1993). We discuss other connections between possibility measures, OCFs, and our logic below, in the context of conditionals.

We now turn our attention to conditional logic. Lewis' main goal in considering preferential structures was to capture a counterfactual conditional $\rightarrow$, where $\psi \rightarrow \varphi$ is read as "if $\psi$ were the case, then $\varphi$ would be true" as in "if kangaroos had no tails, then they would topple over". He takes this to be true at a world $w$ if, in all the worlds "closest" to $w$ (where closeness is defined by a preorder $\succeq$) where kangaroos don't have tails, it is the case that kangaroos topple over.[7]

More abstractly, in the case where $W$ is finite, for a subset $V \subseteq W$, let $\text{best}(V) = \{v \in V : v' \succ v \text{ implies } v' \notin V\}$. Thus, $\text{best}(V)$ consists of all worlds $v \in V$ such that no world $v' \in V$ is considered more likely than $v$. (We take $\text{best}(\emptyset) = \emptyset$.)

If $W$ is finite, we define

$$(M, w) \models \psi \rightarrow \varphi \text{ if } \text{best}([\![\psi]\!]_M) \subseteq [\![\varphi]\!]_M.$$

Thus, $\psi \rightarrow \varphi$ is true exactly if $\varphi$ is true at the most likely (or closest) worlds where $\varphi$ is true.

---

6. Actually, the axiomatization given in (Fariñas del Cerro & Herzig, 1991) is not quite complete as stated; to get completeness, we must replace their axiom QPL4—*true* $\geq \varphi$—by the axiom $\varphi \geq false$ [Luis Fariñas del Cerro, private communication, 1996].

7. To really deal appropriately with counterfactuals, we require not one preorder $\succeq$, but a possibly different preorder $\succeq_w$ for each world $w$, since the notion of closeness in general depends on the actual world. We ignore this issue here, since it is somewhat tangential to our concerns.

For infinite domains, this definition does not quite capture our intentions. For example, in Example 2.2, we have best$(W_\infty) = \emptyset$. It follows that if $M = (W_\infty, \succeq, \pi)$, then $M \models true \rightarrow \neg p$ even if $\pi$ makes $p$ true at every world in $W_\infty$. We certainly would not want to say that "if *true* were the case, then $p$ would be false" is true if $p$ is true at all the worlds in $W_\infty$! The solution here is again a generalization of Lewis's definition in the case of totally ordered worlds, and is much like that for $\succ^s$ in infinite domains. We say $M \models \psi \rightarrow \varphi$ if for all $u \in [\![\neg\varphi \wedge \psi]\!]_M$, there exists a world $v \in [\![\varphi \wedge \psi]\!]_M$ such that $v \succ u$ and $v$ dominates $[\![\neg\varphi \wedge \psi]\!]_M$. This definition agrees with the definition given above for the case of finite $W$.

**Lemma 4.1:** *If $W$ is finite, then* best$([\![\psi]\!]_M) \subseteq [\![\varphi]\!]_M$ *iff for all $u \in [\![\neg\varphi \wedge \psi]\!]_M$, there exists a world $v \in [\![\varphi \wedge \psi]\!]_M$ such that $v \succ u$ and $v$ dominates $[\![\neg\varphi \wedge \psi]\!]_M$.*

**Proof:** See Appendix A. ∎

Lewis (1973) argues that this definition of $\rightarrow$ captures many of our intuitions for counterfactual reasoning. We can give $\rightarrow$ another interpretation, perhaps more natural if we are thinking in terms of likelihood. We often want to say that $\varphi$ is more likely than not—in $\mathcal{L}$, this can be expressed as $\varphi \gg \neg\varphi$. More generally, we might want to say that *relative to $\psi$*, or *conditional on $\psi$ being the case*, $\varphi$ is more likely than not. By this we mean that if we restrict to worlds where $\psi$ is true, $\varphi$ is more likely than not, that is, the worlds where $\varphi \wedge \psi$ is true are more likely than the worlds where $\neg\varphi \wedge \psi$ is true.

Let us define $\psi \rightarrow' \varphi$ to be an abbreviation for $K\neg\psi \vee (\varphi \wedge \psi \gg \neg\varphi \wedge \psi)$. That is, $\psi \rightarrow' \varphi$ is true vacuously in a structure $M$ if $\psi$ does not hold in any world in $M$; otherwise, it holds if $\varphi$ is more likely than not in the worlds satisfying $\psi$.

Although the intuition for $\rightarrow'$ seems, on the surface, quite different from that for $\rightarrow$, especially in finite domains, it is almost immediate from their formal definitions that they are equivalent. (This connection between $\rightarrow$ and $\rightarrow'$ was already observed by Lewis (1973) in the case of total proeorders.)

**Lemma 4.2:** *For all structures $M$, we have $M \models \psi \rightarrow \varphi$ iff $M \models \psi \rightarrow' \varphi$.*

**Proof:** This is almost immediate from the definitions. See Appendix A for details. ∎

Given Lemma 4.2, we can write $\rightarrow$ for both $\rightarrow$ and $\rightarrow'$. The lemma also allows us to apply the known results for conditional logic to the logic of relative likelihood defined here. In particular, the results of (Friedman & Halpern, 1994) show that the validity problem for the logic of Section 3 is co-NP complete, no harder than that of propositional logic, for the case of both partial and total proeorders.[8]

More recently, $\rightarrow$ has been used to capture nonmonotonic default reasoning (Kraus et al., 1990; Boutilier, 1994). In this case, a statement like *Bird $\rightarrow$ Fly* is interpreted as "birds typically fly", or "by default, birds fly". The semantics does not change: *Bird $\rightarrow$ Fly* is true if in the most likely worlds satisfying *Bird*, *Fly* holds as well. Dubois and Prade (1991) have shown that possibility can be used to give semantics to defaults as well, where $\psi \rightarrow \varphi$ is interpreted as $\text{Poss}(\psi) = 0$ or $\text{Poss}(\varphi \wedge \psi) > \text{Poss}(\varphi \wedge \neg\psi)$. Of course, this is just the analogue of the definition of $\rightarrow$ in terms of $\succ^s$. Goldszmidt and Pearl (1992) have shown that a similar approach works if we use Spohn's OCFs.

---

8. We remark that there are also well known axiomatizations for various conditional logics (Burgess, 1981; Friedman & Halpern, 1994; Lewis, 1973). These do not immediately give us a complete axiomatization for the logic of relative likelihood considered here, since we must find axioms in the language with $\gg$, not in the language with $\rightarrow$.

These results are clarified and unified in (Friedman & Halpern, 1997). Suppose we start with some mapping Pl of sets to a partially ordered space with minimal element $\perp$ (such a mapping is called a *plausibility measure* in (Friedman & Halpern, 1997)). Define $\psi \rightarrow \varphi$ as $\text{Pl}(\varphi) = \perp$ or $\text{Pl}(\psi \wedge \varphi) > \text{Pl}(\psi \wedge \neg\varphi)$. Then it is shown that $\rightarrow$ satisfies the *KLM properties*—the properties isolated by Kraus, Lehmann, and Magidor (1990) as forming the core of default reasoning—if and only if Pl is qualitative, at least when restricted to disjoint sets.[9] Since $\succeq^s$, Poss, and OCFs give rise to qualitative orders on $2^W$, it is no surprise that they should all lead to logics that satisfy the KLM properties.

We remark we can also start with $\rightarrow$, and then define $\gg$ in terms of $\rightarrow$. There are, in fact, three related ways of doing so. Define $\varphi \gg' \psi$ to be an abbreviation for $((\varphi \vee \psi) \rightarrow (\varphi \wedge \neg\psi)) \wedge \neg((\varphi \vee \psi) \rightarrow \psi)$; define $\varphi \gg'' \psi$ to be an abbreviation for $\neg(\varphi \rightarrow \psi) \wedge ((\varphi \vee \psi) \rightarrow \neg\psi)$; define $\varphi \gg''' \psi$ to be an abbreviation for $\neg(\varphi \rightarrow \textit{false}) \wedge ((\varphi \vee \psi) \rightarrow (\varphi \wedge \neg\psi))$.

**Proposition 4.3:** *For all structures $M$, the following are equivalent:*

   *(a)* $M \models \varphi \gg \psi$

   *(b)* $M \models \varphi \gg' \psi$

   *(c)* $M \models \varphi \gg'' \psi$.

   *(d)* $M \models \varphi \gg''' \psi$.

The first translation is essentially due to Kraus, Lehmann, and Magidor (1990), the second is essentially due to Freund (1993), and the third is due to Lewis (1973). Since the equivalences are so close to those already in the literature, we omit the proof of this result here. Using these equivalences and results of (Kraus et al., 1990), Daniel Lehmann [private correspondence, 1996] has provided an alternate proof for Theorem 3.1. See the remarks after the proof of that theorem in Appendix A for a few more details.

## 5. Conclusion

We have investigated a notion of relative likelihood starting with a preferential ordering on worlds. This notion was earlier studied by Lewis (1973) in the case where the preferential order is a total preorder; the focus of this paper is on the case where the preferential order is a partial preorder. Our results show that there are significant differences between the totally ordered and partially ordered case. By focusing on the partially ordered case, we bring out the key role of the qualitative property (Axiom L3), whose connections to conditional logic were already observed in (Friedman & Halpern, 1997).

---

9. That is, if $V_1$, $V_2$, and $V_3$ are *disjoint* sets, we require that if $\text{Pl}(V_1 \cup V_2) > \text{Pl}(V_3)$ and $\text{Pl}(V_1 \cup V_3) > \text{Pl}(V_2)$, then $\text{Pl}(V_1) > \text{Pl}(V_2 \cup V_3)$. The result also requires the assumption that if $\text{Pl}(U) = \text{Pl}(V) = \perp$, then $\text{Pl}(U \cup V) = \perp$.

## Acknowledgements

Many interesting and useful discussions on plausibility with Nir Friedman formed the basis for this paper; Nir also pointed out the reference (Doyle et al., 1991). Daniel Lehmann, Emil Weydert, and the referees of the paper also provided useful comments. A preliminary version of this paper appears in *Uncertainty in Artificial Intelligence, Proceedings of the Twelfth Conference*, 1996, edited by E. Horvitz and F. Jensen. Most of this work was carried out while the author was at the IBM Almaden Research Center. IBM's support is gratefully acknowledged. The work was also supported in part by NSF under grants IRI-93-03109 and IRI-96-25901, and by the Air Force Office of Scientific Research under contract F49620-96-1-0323.

## Appendix A. Proofs

We repeat the statements of the results we are proving here for the convenience of the reader.

**Lemma 2.4:** *If $\rhd$ is an orderly qualitative relation on $2^W$, then $\rhd$ is transitive and satisfies the union property.*

**Proof:** Suppose $\rhd$ is an orderly qualitative relation. To see that $\rhd$ is transitive, suppose $V_1 \rhd V_2$ and $V_2 \rhd V_3$. Since $\rhd$ is orderly, it follows that $(V_1 \cup V_3) \rhd V_2$ and $(V_1 \cup V_2) \rhd V_3$. Since $\rhd$ is qualitative, it follows that $V_1 \rhd (V_2 \cup V_3)$. From the fact that $\rhd$ is orderly, we get that $V_1 \rhd V_3$. Thus, $\rhd$ is transitive, as desired.

To see that $\rhd$ satisfies the union property, suppose $V_1 \rhd V_2$ and $V_1 \rhd V_3$. Since $\rhd$ is orderly, we have that $(V_1 \cup V_3) \rhd V_2$ and $(V_1 \cup V_2) \rhd V_3$. Using the fact that $\rhd$ is qualitative, we get that $V_1 \rhd (V_2 \cup V_3)$. Hence, $\rhd$ satisfies the union property. ■

**Proposition 2.5:**

(a) *If $\succeq$ is a partial preorder on $W$, then $\succeq^s$ is an orderly partial preorder on $2^W$ that satisfies the union property.*

(b) *If $\succeq$ is a partial preorder on $W$, then $\succ'$ is an orderly strict partial order on $2^W$.*

(c) *If $\succ$ is a strict partial order on $W$, then $\succ^s$ is an orderly qualitative strict partial order on $2^W$.*

**Proof:** We prove part (c) here; the proof of parts (a) and (b) is similar in spirit, and is left to the reader. The fact that $\succ^s$ is an orderly strict partial order is straightforward, and is also left to the reader. To see that $\succ^s$ is qualitative, suppose $V_1 \cup V_2 \succ^s V_3$ and $V_1 \cup V_3 \succ^s V_2$. Let $v \in V_2 \cup V_3$. We must show that there is some $v' \in V_1$ that dominates $V_2 \cup V_3$ such that $v' \succ v$. Suppose without loss of generality that $v \in V_2$ (an identical argument works if $v \in V_3$). Since $V_1 \cup V_3 \succ^s V_2$, there is some $u \in V_1 \cup V_3$ that dominates $V_2$ such that $u \succ v$. If $u$ dominates $V_3$, then it clearly dominates $V_2 \cup V_3$ and it must be in $V_1$, so we are done. Thus, we can assume that $u$ does not dominate $V_3$, so there is some element $u' \in V_3$ such that $u' \succ u$. Since $V_1 \cup V_2 \succ^s V_3$, there must be some $v' \in V_1 \cup V_2$ such that $v'$ dominates $V_3$ and $v' \succ u'$. Since $u$ dominates $V_2$ and $u' \succ u$, it follows that $u$ dominates $V_2$. Since $v' \succ u'$, we have that $v'$ dominates $V_2$. Hence, $v'$ dominates $V_2 \cup V_3$. It follows that $v'$ cannot be in $V_2$, so it must be in $V_1$. Thus, we have an element in $V_1$, namely $v'$, such that $v' \succ v$ and $v'$ dominates $V_2 \cup V_3$, as desired. ∎

**Lemma 2.6:** *If $\succeq$ is a total preorder, then the strict partial order $\succ$ determined by $\succeq$ is modular. Moreover, if $\rhd$ is a modular, strict partial order on $W$, then there is a total preorder $\succeq$ on $W$ such that $\rhd$ is the strict partial order determined by $\succeq$.*

**Proof:** Suppose $\succeq$ is a total preorder. To see that $\succ$ is modular, suppose that $w_1 \succ w_2$. Given an arbitrary $w_3$, if $w_3 \succeq w_1$, it follows from the transitivity of $\succeq$ that $w_3 \succ w_2$. On the other hand, if it is not the case that $w_3 \succeq w_1$, then $w_1 \succ w_3$. Thus, we have that either $w_3 \succ w_2$ or $w_1 \succ w_3$, so $\succ$ is modular.

Now suppose that $\rhd$ is a modular strict partial order on $W$. Define $\succeq$ so that $w \succeq v$ either if $w \rhd v$ or if neither $w \rhd v$ nor $v \rhd w$ hold. Clearly, $\succeq$ is reflexive. To see that it is transitive, suppose that $v_1 \succeq v_2$ and $v_2 \succeq v_3$. There are three cases: (1) If $v_1 \rhd v_2$, then since $\rhd$ is modular, we have that either $v_1 \rhd v_3$ or $v_3 \rhd v_2$. We cannot have $v_3 \rhd v_2$, for then we would not have $v_2 \succeq v_3$. Thus, we must have $v_1 \rhd v_3$, and hence $v_1 \succeq v_3$. (2) If $v_2 \rhd v_3$, then using modularity again, we get that either $v_1 \rhd v_3$ or $v_2 \rhd v_1$. Again, we cannot have $v_2 \rhd v_1$, so we must have $v_1 \rhd v_3$, and so we also have $v_1 \succeq v_3$. (3) If neither $v_1 \rhd v_2$ nor $v_2 \rhd v_3$ hold, then we claim that neither $v_1 \rhd v_3$ nor $v_3 \rhd v_1$ hold. For if $v_1 \rhd v_3$, then by modularity, we must have either $v_1 \rhd v_2$ or $v_2 \rhd v_3$. And if $v_3 \rhd v_1$, then either $v_3 \rhd v_2$ or $v_2 \rhd v_1$, which contradicts the assumption that $v_1 \succeq v_2$ and $v_2 \succeq v_3$. Thus, we can again conclude that $v_1 \succeq v_3$. Thus, $\succeq$ is transitive. Finally, it is almost immediate from the definition that $\rhd$ is the strict partial order determined by $\succeq$. ∎

**Lemma 2.7:** *If $\succ$ is a modular relation on $W$, then $\succ^s$ is a modular relation on $2^W$.*

**Proof:** Suppose $\succ$ is modular. We want to show that $\succ^s$ is modular. So suppose that $V_1 \succ^s V_2$, and it is not the case that $V_1 \succ^s V_3$. We must show that $V_3 \succ^s V_2$. Since it is not the case that $V_1 \succ^s V_3$, there must be some $v^* \in V_3$ such that for all $u \in V_1$, we do not have $u \succ v^*$. Now suppose $v \in V_2$. We claim that $v^* \succ v$. To see this, note that since $V_1 \succ^s V_2$, there must be some $u^* \in V_1$ such that $u^* \succ v$. Since $\succ$ is modular, we have that either $u^* \succ v^*$ or $v^* \succ v$. Since, by choice of $v^*$, we do not have $u^* \succ v^*$, we must have $v^* \succ v$. It follows that $V_3 \succ^s V_2$. ∎

**Lemma 2.8:** *If $\rhd$ is a modular strict partial order and satisfies the union property, then $\rhd$ is qualitative.*

**Proof:** Suppose that $\rhd$ is modular strict partial order that satisfies the union property. To see that $\rhd$ is qualitative, suppose that $(V_1 \cup V_2) \rhd V_3$ and $(V_1 \cup V_3) \rhd V_2$. Since $\rhd$ is modular, it follows that either $(V_1 \cup V_2) \rhd V_1$ or $V_1 \rhd V_3$. If $(V_1 \cup V_2) \rhd V_1$, then, using the fact that $\rhd$ satisfies the union property and $(V_1 \cup V_2) \rhd V_3$, we get that $(V_1 \cup V_2) \rhd (V_1 \cup V_3)$. Using transitivity, it follows that $(V_1 \cup V_2) \rhd V_2$. Using the union property again, we get that $(V_1 \cup V_2) \rhd (V_1 \cup V_2)$. This contradicts the assumption that $\rhd$ is irreflexive. Thus, we must have that $V_1 \vartriangleright V_3$. A similar argument shows that $V_1 \vartriangleright V_2$. Using the union property, we get that $V_1 \vartriangleright (V_2 \cup V_3)$, as desired. ∎

**Lemma 2.9:** *If $\succeq$ is a total preorder, then $\succ^s$ and $\succ'$ agree. In general, $U \succ^s V$ implies $U \succ' V$, but the converse does not hold.*

**Proof:** It is immediate from the definitions that $U \succ^s V$ implies $U \succ' V$, and the fact that the converse does not hold is shown by Example 2.1. To show that $\succ^s$ and $\succ'$ are equivalent if $\succeq$ is a total preorder, suppose $U \succ' V$. Clearly $U$ is nonempty, since $V \succeq^s \emptyset$ for all $V$. We want to show that $U \succ^s V$, so we must show that for all $v \in V$, there is some $u \in U$ that dominates $V$ such that $u \succ v$. We actually show that there is some $u \in U$ such that $u \succ v'$ for all $v' \in V$. Suppose not. Then for every $u \in U$, there is some $v_u$ such that we do not have $u \succ v_u$. Since $\succeq$ is a total order, this means that $v_u \succeq u$. But this, in turn, means that $V \succeq^s U$, contradicting our assumption that $U \succ^s V$. Since there is a $u \in U$ such that $u \succ v$ for all $v \in V$, it easily follows that $u^*$ dominates $V$ and that $U \succ^s V$, as desired. ∎

**Theorem 2.10:** *Let $\mathcal{F}$ be a finite algebra of subsets of $W$ (that is, $\mathcal{F}$ is a set of subsets of $W$ that is closed under union and complementation and contains $W$ itself) and let $\vartriangleright$ be an orderly qualitative relation on $\mathcal{F}$.*

(a) *If $\vartriangleright$ is a total preorder on $\mathcal{F}$, then there is a total preorder $\succeq$ on $W$ such that $\vartriangleright$ and $\succeq^s$ agree on $\mathcal{F}$ (that is, for $U, V \in \mathcal{F}$, we have $U \vartriangleright V$ iff $U \succeq^s V$).*

(b) *If $\vartriangleright$ is a strict partial order and each nonempty set in $\mathcal{F}$ has at least $2^{\log(|\mathcal{F}|)^{\log(|\mathcal{F}|)}}$ elements, then there is a partial preorder $\succeq$ on $W$ such that $\vartriangleright$ and $\succ^s$ agree on $\mathcal{F}$.*

**Proof:** Part (a) was already proved in the main text, so we prove part (b) here.

We proceed as follows. We say that a pair $(A, X)$ is *minimal pair* of $\vartriangleright$ if $X \vartriangleright A$ and there is no $X' \subset X$ ("$\subset$" is used here to denote strict subset) such that $X' \vartriangleright A$. The minimal pairs in an ordered relation determine it. Indeed, the following stronger result holds.

**Lemma A.1:** *If $\vartriangleright$ and $\vartriangleright'$ are two orderly qualitative relations on $\mathcal{F}$ such that for each minimal pair $(A, X)$ of $\vartriangleright$ we have $X \vartriangleright A$, and for each minimal pair $(A', X')$ of $\vartriangleright'$ we have $X' \vartriangleright' A'$, then $\vartriangleright = \vartriangleright'$.*

**Proof:** Suppose the assumptions of the lemma holds and that $X \vartriangleright Y$; we must show that $X \vartriangleright' Y$. Suppose $A$ is an atom such that $A \subseteq Y$. Since $\vartriangleright$ is orderly, we must have $X \vartriangleright A$. Let $X'$ be a minimal subset of $X$ such that $X' \vartriangleright A$. Since $\mathcal{F}$ is finite, such an $X'$ must exist. Thus, $(A, X')$ is a minimal pair of $\vartriangleright$. By assumption, we have $X' \vartriangleright' A$. Since $\vartriangleright'$ is orderly, we also have $X \vartriangleright' A$. Thus, we have $X \vartriangleright' A$ for each atom $A \subseteq Y$. Since $\vartriangleright'$ is qualitative, it also satisfies the union property, and hence $X \vartriangleright' Y$ as desired. By symmetry, it follows that $\vartriangleright = \vartriangleright'$. ∎

A *minimal-pair tree* for $\vartriangleright$ is a rooted tree whose nodes are labeled by atoms such that the following conditions are satisfied:

1. If a node in the tree labeled $B$ is an immediate successor of a node labeled $A$, then there must be a minimal pair $(A, X)$ of $\rhd$ such that $B \subseteq X$.

2. If there is a node $t$ in the tree labeled $A$ and a minimal pair of $\rhd$ of the form $(A, X)$, then there must be some atom $B \subseteq X$ such that a node labeled $B$ is an immediate successor of $t$.

3. There does not exist a path in the tree such that two nodes in the path have the same label.

4. A node does not have two distinct successors with the same label.

A minimal-pair tree is *rooted at $A$* if the root of the tree is labeled by the atom $A$.

Since a subset of $\mathcal{F}$ can be written in a unique way as the union of atoms, there is a 1-1 correspondence between subsets of $\mathcal{F}$ and sets of atoms. Thus, there are exactly $\log(|\mathcal{F}|)$ atoms. Because of the third condition, a path in the tree can have length at most $\log(|\mathcal{F}|)$. Since all the atoms on the path must be distinct, it follows that there are at most $\log(|\mathcal{F}|)! \leq \log(|\mathcal{F}|)^{\log(|\mathcal{F}|)} (= |\mathcal{F}|^{\log\log(|\mathcal{F}|)})$ possible paths in a tree rooted at $A$. We can identify a tree with the set of its paths, which means that there are at most $2^{\log(|\mathcal{F}|)!} \leq 2^{\log(|\mathcal{F}|)^{\log(|\mathcal{F}|)}}$ possible trees rooted at an atom $A$.

We now label each element in $W$ with a minimal-pair tree in such a way that every element of atom $A$ is labeled by a tree rooted at $A$, and every minimal-pair tree rooted at $A$ is the label of some element of $A$. Since we have assumed that $A$ has at least $2^{\log(|\mathcal{F}|)^{\log(|\mathcal{F}|)}}$ elements, there is such a labeling. Let $L(w)$ be the label of node $w$. We define $\succ$ on $W$ so that $w' \succ w$ iff $L(w')$ is a proper subtree of $L(w)$. Clearly $\succ$ is a strict partial order.

We claim that $\rhd$ agrees with $\succ^s$. By Lemma A.1, it suffices to show that if $(A, X)$ is a minimal pair of $\rhd$, then $X \succ^s A$, and if $(A, X)$ is a minimal pair of $\succ^s$, then $X \rhd A$. So suppose $(A, X)$ is a minimal pair of $\rhd$. We want to show that $X \succ^s A$. Let $w \in A$, and suppose that $L(w) = T$. Thus, $T$ is a minimal-pair tree rooted at $A$. The construction of minimal-pair trees guarantees that there is a successor of the root in tree $T$ labeled $B$ for some atom $B \subseteq X$. Consider the subtree of $T$ rooted at $B$. This is a minimal-pair tree rooted at $B$, and hence must be the label of some $w' \in B \subseteq X$. Thus, $w' \succ w$. It follows that $X \succ^s A$.

Now suppose that $(A, X)$ is a minimal pair of $\succ^s$. We want to show that $X \rhd A$. Suppose not. As we show below, this means that there exists a minimal-pair tree $T$ rooted at $A$ such that no node in $T$ is labeled by an atom contained in $X$. Let $w$ be an element of $A$ with $L(w) = T$. By construction, there is no element $w'$ of $X$ such that $w' \succ w$. Thus, we do not have $X \succ^s A$, contradicting our initial assumption.

It remains to show that there exists a minimal-pair tree $T$ rooted at $A$ such that no node in $T$ is labeled by an atom contained in $X$. Clearly, we cannot have $X' \rhd A$ for some $X' \subset X$, for then, by the preceding argument, we would have $X' \succ^s A$, and $(A, X)$ would not be a minimal pair of $\succ^s$. It follows that if $(A, Y)$ is a minimal pair of $\rhd$, then we must have $Y - X \neq \emptyset$.

Two general fact about orderly, qualitative relations will be useful in our construction:

**Lemma A.2:** *If $\rhd'$ is a qualitative relation on $\mathcal{F}$ and $Y \rhd' X$, then $(Y - X) \rhd' X$.*

**Proof:** Notice that $(Y - X) \cup X \; \rhd' \; X$. Since $\rhd'$ is qualitative, it follows that $(Y - X) \; \rhd' \; X$ (take both $V_2$ and $V_3$ in the definition of qualitative to be $X$).

**Lemma A.3:** *If* $\rhd'$ *is an orderly, qualitative relation such that* $(X_1 \cup X_2) \; \rhd' \; X_3$ *and* $X' \; \rhd' \; X_2$, *then* $(X_1 \cup X') \; \rhd' \; X_3$.

**Proof:** Since $\rhd'$ is orderly, our assumptions imply that $(X_1 \cup X' \cup X_2) \; \rhd' \; X_3$ and that $(X_1 \cup X' \cup X_3) \; \rhd' \; X_2$. Since $\rhd'$ is qualitative, it follows that $(X_1 \cup X') \; \rhd \; (X_2 \cup X_3)$. The result follows using the fact that $\rhd'$ is orderly again. ∎

We start by constructing a tree whose nodes are labeled by atoms and whose root is labeled by $A$. We proceed in $\log(|\mathcal{F}|) + 1$ stages. At each stage, we have a tree whose nodes are labeled by atoms. At stage 0, we just take a single node labeled by $A$. Suppose we have constructed a tree whose nodes are labeled by atoms and whose root is labeled by $A$ at stage $k < \log(|\mathcal{F}|)$. For stage $k + 1$, for each leaf $t$ in the stage-$k$ tree, if $t$ is labeled by $B$, then for each atom $C$, if there is a minimal pair $(B, Y)$ of $\rhd$ with $C \subseteq Y$, we add a successor to $t$ labeled $C$. We call the tree constructed at the end of stage $\log(|\mathcal{F}|)$ the *full tree for* $A$.

We next mark nodes in the full tree in stages. At the $k$th stage, for each atom $B$, we mark an unmarked node $t$ labeled $B$ if one of the following three conditions holds: (1) $B \subseteq X$, (2) there is an ancestor of $t$ in the tree also labeled $B$, (3) there is a minimal pair $(B, Y)$ of $\rhd$ and all the successors of $t$ with a label contained in $Y$ were marked at an earlier stage. If there are no unmarked nodes satisfying one of these three conditions at stage $k$, then the marking process stops. Otherwise, we continue to stage $k + 1$. Since there are only finitely many nodes, this marking process is guaranteed to terminate.

Our goal is to show that, at the end of the marking process, the root of the full tree is unmarked. For if this is the case, let $T$ be the subtree of the full tree consisting of all the unmarked nodes all of whose ancestors are unmarked. It is easy to check that $T$ is a minimal-pair tree and our marking procedure guarantees that no node in $T$ is labeled by an atom contained in $X$, so we are done.

To see that the root of the full tree is unmarked, we proceed as follows: Define a *0-cover* for a node $t$ in the full tree to be just $t$ itself. Suppose we have defined a $k$-cover for $t$. A set $Z$ of nodes is a $(k+1)$-*cover* for $t$ if there exists a $k$-cover $Z'$ for $t$ such that for some node $t'$ in $Z'$ labeled $B$ and some minimal pair $(B, Y)$ of $\rhd$, we have that $Z$ consists of all the nodes in $Z'$ except for $t'$, together with all the successors of $t'$ that are labeled by an atom contained in $Y$. An easy argument by induction on $k$ shows the following.

**Lemma A.4:** *If* $Z$ *is a $k$-cover for a node* $t$ *labeled* $C$ *and* $k > 0$, *then there exist a set* $Y$ *such that* $(C, Y)$ *is a minimal pair of* $\rhd$, *successors* $t_1, \ldots, t_m$ *of* $t$ *in the full tree, atoms* $D_1, \ldots, D_m$ *such that* $Y = \cup_{i=1}^m D_i$ *and* $D_i$ *is the label of* $t_i$, $i = 1, \ldots, m$, *and a partition* $Z_1, \ldots, Z_m$ *of* $Z$ *into disjoint subsets such that* $Z_i$ *is a $k_i$-cover for* $t_i$, $i = 1, \ldots, m$, *for some* $k_i < k$.

Given a set $Z$, let $U_Z^n$ consist of the union of the atoms labeling the nodes of $Z$ still unmarked at the $n$th stage (we take $U_Z^0$ to be the union of the atoms labeling the nodes in $Z$); given a node $t$, let $V_t$ consist of the union of the atoms $D$ labeling ancestors of $t$ such that $t$ or some descendent of $t$ has the label $D$.

The key fact is the following result.

**Lemma A.5:** *If $Z$ is a $k$-cover for a node $t$ labeled $C$ and $k > 0$, then $(U_Z^n \cup V_t \cup X) \rhd C$.*

**Proof:** We proceed by induction on $k$, with a subinduction on $n$. If $k = 1$, then there is some set $Y$ such that $(C, Y)$ is a minimal pair of $\rhd$ and the nodes in $Z$ are labeled by the atoms contained in $Y$. Since $(C, Y)$ is a minimal pair of $\rhd$, we have that $Y \rhd C$. Since $\rhd$ is orderly, $(Y \cup V_t \cup X) \rhd C$. By definition, $U_Z^0 = Y$, so this takes care of the case $n = 0$. Suppose $n > 0$ and $(U_Z^{n-1} \cup V_t \cup X) \rhd C$. For the inductive step, it suffices to show that that if

$$(Y' \cup V_t \cup X) \rhd C \tag{1}$$

and $D$ is the label of a node $t'$ in $Z$ marked at stage $n$, then

$$((Y' - D) \cup V_t \cup X) \rhd C. \tag{2}$$

So suppose that (1) holds and $D$ is the label of $t'$. We must consider how $t'$ was marked. If $D \subseteq X$ then (2) is immediate. If there is an ancestor of $t'$ also labeled $D$, then $D \subseteq V_{t'}$, and, since $k = 1$, we have $V_{t'} \subseteq V_t \cup C$. Thus, since $\rhd$ is orderly, it follows that

$$((Y' - D) \cup V_t \cup X \cup C) \rhd C, \tag{3}$$

so (2) follows from Lemma A.2. Finally, suppose there is some minimal pair $(D, Y'')$ of $\rhd$ such that all the successors of $t'$ labeled by an atom in $Y''$ are marked by stage $n - 1$. Let $Z'$ consist of the successors of $t'$ labeled by atoms contained in $Y''$. $Z'$ is a 1-cover for $t'$. Since $U_{Z'}^{n-1} = \emptyset$, it follows from the induction hypothesis that $(V_{t'} \cup X) \rhd D$. Again, since $V_{t'} \subseteq V_t \cup C$, (3) follows orderliness and Lemma A.3, and the desired (2) follows from Lemma A.2.

Now suppose $k > 1$. Let $Y, D_1, \ldots, D_m, Z_1, \ldots, Z_m, t_1, \ldots, t_m$ be the sets and nodes guaranteed to exist by Lemma A.4. Since $Z_i$ is a $k_i$-cover for $t_i$ for some $k_i < k$, by the induction hypothesis, $(U_{Z_i}^n \cup V_{t_i} \cup X) \rhd D_i$. Since $U_{Z_i}^n \subset U_Z^n$ and $V_{t_i} \subseteq V_t \cup C$, by orderliness, we have that $(U_Z^n \cup V_t \cup X \cup C) \rhd D_i$. Since $\rhd$ is qualitative, we have that $(U_Z^n \cup V_t \cup X \cup C) \rhd (\cup_i D_i)$. Since $\cup_i D_i = Y$ and $Y \rhd C$, (3) now follows, and again (2) follows from Lemma A.2. ■

Finally, suppose, by way of contradiction, that the root $r$ of the full tree is marked, say at stage $n$ of the marking process. Lemma A.2 assures us that $A \cap X = \emptyset$, since $(A, X)$ is a minimal pair of $\succ^s$, so condition (1) of the marking process does not apply. Since $r$ has no ancestors, condition (2) does not apply either. Thus, there must be some minimal pair $(A, Y)$ of $\rhd$ such that all the nodes in the set $Z$ consisting of the successors of $r$ in the full tree that are labeled by atoms contained in $Y$ are marked by stage $n - 1$. Thus, we $U_Z^{n-1} = \emptyset$. Since $V_r = \emptyset$ and $Z$ is a 1-cover for $r$, by Lemma A.5, it follows that $X \rhd A$, contradicting our original assumption. ■

For the completeness theorems (Theorems 3.1 and 3.2) it is convenient to start by proving Theorem 3.2, since it is simpler, and contains the key ideas for the proof of Theorem 3.1.

**Theorem 3.2:** *$AX^M$ is a sound and complete axiomatization of the language $\mathcal{L}$ with respect to totally preordered preferential structures.*

**Proof:** Soundness was proved in the main text, so we just consider completeness here. Suppose that $\varphi$ is consistent with $AX^M$. We want to show that that $\varphi$ is satisfiable in a totally preordered preferential structure. Let $\varphi_1, \ldots, \varphi_m$ be the basic likelihood formulas that are subformulas of $\varphi$. By definition, $\varphi$ is a Boolean combination of these formulas. Define an *atom over $\varphi$* to be a conjunction of the form $\psi_1 \wedge \ldots \wedge \psi_m$, where $\psi_i$ is either $\varphi_i$ or $\neg \varphi_i$. Using straightforward propositional reasoning (L1 and MP), it is straightforward to show that $\varphi$ is provably equivalent to the disjunction of the consistent atoms over $\varphi$. Thus, since $\varphi$ is consistent, some atom over $\varphi$, say $\sigma$, is consistent. We now construct a totally ordered preferential structure satisfying $\sigma$. Clearly this structure will satisfy $\varphi$ as well.

Let $p_1, \ldots, p_n$ be the primitive propositions in $\Phi$ that appear in $\varphi$. Let $\Sigma$ consist of all the $N = 2^n$ truth assignments to these primitive propositions. We take $W = \Sigma$ and let $\mathcal{F}$ consist of all subsets of $\Sigma$. We define a total preorder $\triangleright$ on $\mathcal{F}$ as follows. Notice that to each set $V$ in $\mathcal{F}$, there corresponds a propositional formula $\varphi_V$ that is made true precisely by the truth assignments in the subset $V'$ of $\Sigma$ that corresponds to $V$. To be precise, given a truth assignment $\alpha$, let $\varphi_\alpha$ consist of the conjunction $q_1 \wedge \ldots \wedge q_n$, where $q_i$ is $p_i$ if $\alpha(p_i) = \mathbf{true}$, and $\neg p_i$ otherwise. Let $\varphi_V$ be the disjunction of the formulas $\varphi_\alpha$ for $\alpha \in V'$. (We take the empty disjunction to be the formula *false*.) Notice for future reference that $\varphi_{V_1 \cup V_2}$ is provably equivalent to $\varphi_{V_1} \vee \varphi_{V_2}$. Conversely, for every propositional formula $\psi$ that mentions only the primitive propositions in $\{p_1, \ldots, p_n\}$, there is a corresponding subset $A_\psi$ in $\mathcal{F}$ that consists of all truth assignments that make $\psi$ true.

We define a binary relation $\triangleright$ on $\mathcal{F}$ as follows: $V \triangleright V'$ iff $AX \vdash \sigma \Rightarrow (\varphi_V \gg \varphi_{V'})$. We claim that $\triangleright$ is a modular, qualitative, strict partial order on $\mathcal{F}$. The fact that $\triangleright$ is irreflexive follows easily from L2; the fact that it is orderly follows from L4; the fact that it is qualitative follows from L3; transitivity follows from the fact that $\triangleright$ is qualitative and orderly, by Lemma 2.4; modularity follows from L5. All these arguments are straightforward. We prove the fact that $\triangleright$ is qualitative here, and leave the remaining arguments to the reader.

Suppose $V_1, V_2, V_3 \in \mathcal{F}$, $(V_1 \cup V_2) \triangleright V_3$, and $(V_1 \cup V_3) \triangleright V_2$. Our assumptions and the definition of $F$ imply that $AX \vdash \sigma \Rightarrow ((\varphi_{V_1} \vee \varphi_{V_2}) \gg \varphi_{V_3})$ and that $AX \vdash \sigma \Rightarrow ((\varphi_{V_1} \vee \varphi_{V_3}) \gg \varphi_{V_2})$. By L3 and straightforward propositional reasoning, we get that $AX \vdash \sigma \Rightarrow (\varphi_{V_1} \gg (\varphi_{V_2} \vee \varphi_{V_3}))$, so that $V_1 \triangleright (V_2 \cup V_3)$, as desired.

By Lemma 2.6, there is a total preorder $\triangleright'$ on $\mathcal{F}$ such that $\triangleright$ is the strict partial order determined by $\triangleright'$. By Theorem 2.10(a), there is total preorder $\succeq$ on $W$ such that $\triangleright'$ and $\succeq^s$ agree on $\mathcal{F}$. Since $\succ'$ is the strict partial order determined by $\succeq^s$, it follows that $\succ'$ and $\triangleright$ agree. Since $\succeq$ is a total preorder, by Lemma 2.9, $\succ^s$ and $\triangleright$ agree. Let $M = (W, \succeq, \pi)$. We now claim that for each formula $\varphi_j$ that is one of the basic likelihood formulas that is a subformula of $\varphi$, we $M \models \varphi_i$ iff $\varphi_j$ is a conjunct of $\sigma$. For suppose $\varphi_j$ is of the form $\psi \gg \psi'$. If $\varphi_j$ is a conjunct of $\sigma$, then clearly $AX \vdash \sigma \Rightarrow (\psi \gg \psi')$. Thus, $A_\psi \triangleright A_{\psi'}$ by definition, so $A_\psi \succ^s A_{\psi'}$ by construction. Since $A_\psi$ and $A_{\psi'}$ consist of the worlds in $W$ where $\psi$ and $\psi'$, respectively, are true, it follows that $(M, w) \models \psi \gg \psi'$. On the other hand, if $\neg \varphi_j$ is a conjunct of $\sigma$, then we have $AX \vdash \sigma \Rightarrow \neg(\psi \gg \psi')$. We must have $(M, w) \models \neg(\psi \gg \psi')$, for if $(M, w) \models \psi \gg \psi'$, the same arguments as those above would imply that $A_\psi \triangleright A_{\psi'}$, and so $AX \vdash \sigma \Rightarrow \psi \gg \psi'$, contradicting the consistency of $\sigma$.

Thus, $M$ satisfies $\sigma$, and hence $\varphi$. ∎

**Theorem 3.1:** *AX is a sound and complete axiomatization of the language $\mathcal{L}$ with respect to preferential structures.*

**Proof:** Again, soundness is proved in the main text, so we just consider completeness. The ideas are much in the spirit of the proof of Theorem 3.2. We again take $\Sigma$ to consist of all the $N = 2^n$ truth assignments to these primitive propositions. However, since we plan to apply part (b) of Theorem 2.10, we no longer take $W = \Sigma$. Rather, we take $W$ to consist of $2^{n^n}$ copies of the truth assignments in $\Sigma$. More precisely, let $W$ consist of the $N 2^{n^n}$ worlds of the form $w_\alpha^i$, such that $i = 1, \ldots, 2^{n^n}$ and $\alpha \in \Sigma$. Let $\mathcal{F}$ consist of all subsets of $W$ that correspond to subsets of $\Sigma$; that is, $V \in \mathcal{F}$ iff there exists some $V' \subseteq \Sigma$ such that $V = \{w_\alpha^i : i = 1, \ldots, 2^{n^n}, \alpha \in V'\}$. Clearly, $\mathcal{F}$ is a finite algebra with $N$ elements, and each nonempty set in $\mathcal{F}$ has at least $2^{\log(|\mathcal{F}|)^{\log(|\mathcal{F}|)}}$ elements.

We define a strict partial order $\triangleright$ on $\mathcal{F}$ just as in the proof of Theorem 3.1: $V \triangleright V'$ iff $AX \vdash \sigma \Rightarrow \varphi_V \gg \varphi_{V'}$. As before, $\triangleright$ is an orderly qualitative strict partial order on $\mathcal{F}$. It is not necessarily modular, since we no longer have L5.

By Theorem 2.10(b), there is a partial preorder $\succeq$ on $W$ such that $\triangleright$ and $\succ^s$ agree on $\mathcal{F}$. Let $M = (W, \succeq, \pi)$, where $\pi(w_\alpha^i) = \alpha$. Just as in the proof of Theorem 3.2, we can now show that $M \models \sigma$. ∎

As we noted earlier, Daniel Lehmann has found another proof for Theorem 3.1, using results from (Kraus et al., 1990). To show that a formula in $\mathcal{L}$ that is consistent with AX is satisfiable, he first translates it to a formula of conditional logic (using Proposition 4.3). It then follows from the representation theorem of (Kraus et al., 1990) that this translated formula is satisfiable in a preferential structure. The original formula is then satisfiable in the same structure. This proof allows us to avoid using Theorem 2.10 altogether. However, we feel that Theorem 2.10 gives insight into the connection between partial orders on worlds and partial orders on sets of worlds, and thus is of interest in its own right.

**Proposition 3.4:** *If a formula is satisfiable in a totally preordered preferential structure, then it is satisfiable in a totally preordered preferential structure with at most one world per truth assignment.*

**Proof:** This follows immediately from the completeness proof of Theorem 3.2; the structure constructed in that proof has one world per truth assignment. ∎

**Lemma 4.1:** *If $W$ is finite, then $\text{best}(\llbracket \psi \rrbracket_M) \subseteq \llbracket \varphi \rrbracket_M$ iff for all $u \in \llbracket \neg\varphi \wedge \psi \rrbracket_M$, there exists a world $v \in \llbracket \varphi \wedge \psi \rrbracket_M$ such that $v \succ u$ and $v$ dominates $\llbracket \neg\varphi \wedge \psi \rrbracket_M$.*

**Proof:** Suppose $\text{best}(\llbracket \psi \rrbracket_M) \subseteq \llbracket \varphi \rrbracket_M$. If $u \in \llbracket \neg\varphi \wedge \psi \rrbracket_M$, then we cannot have $u \in \text{best}(\llbracket \psi \rrbracket_M)$, since $\text{best}(\llbracket \psi \rrbracket_M) \subseteq \llbracket \varphi \rrbracket_M$. Since $W$ is finite, there exists a world $v \in \text{best}(\llbracket \psi \rrbracket_M)$ such that $v \succ u$. Since $v \in \text{best}(\llbracket \psi \rrbracket_M)$, we have that $v$ dominates $\llbracket \neg\varphi \wedge \psi \rrbracket_M$. Conversely, suppose that for all $u \in \llbracket \neg\varphi \wedge \psi \rrbracket_M$, there exists a world $v \in \llbracket \varphi \wedge \psi \rrbracket_M$ such that $v \succ u$ and $v$ dominates $\llbracket \neg\varphi \wedge \psi \rrbracket_M$. If $u \in \text{best}(\llbracket \psi \rrbracket_M)$, then we must have $u \in \llbracket \varphi \rrbracket_M$, for if not, there exists a world $v \in \llbracket \varphi \wedge \psi \rrbracket_M$ such that $v \succ u$, contradicting the assumption that $u \in \text{best}(\llbracket \varphi \rrbracket_M)$. ∎

**Lemma 4.2:** *For all structures $M$, we have $M \models \psi \to \varphi$ iff $M \models \psi \to' \varphi$.*

**Proof:** Suppose $M \models \psi \to \varphi$. By definition, this means that for all $u \in [\![\neg\varphi \wedge \psi]\!]_M$, there exists a world $v \in [\![\varphi \wedge \psi]\!]_M$ such that $v \succ u$ and $v$ dominates $[\![\neg\varphi \wedge \psi]\!]_M$. It immediately follows that (a) if $[\![\varphi \wedge \psi]\!]_M = \emptyset$ then $[\![\neg\varphi \wedge \psi]\!]_M = \emptyset$ and (b) if $[\![\varphi \wedge \psi]\!]_M \neq \emptyset$, then by definition we have $M \models \varphi \wedge \psi \gg \neg\varphi \wedge \psi$. It follows from (a) that if $[\![\varphi \wedge \psi]\!]_M = \emptyset$ then $[\![\psi]\!]_M = \emptyset$, so $M \models K\neg\psi$, and from (b) that if $[\![\varphi \wedge \psi]\!]_M \neq \emptyset$ then $M \models \varphi \wedge \psi \gg \neg\varphi \wedge \psi$. Thus, $M \models K\neg\psi \vee (\varphi \wedge \psi \gg \neg\varphi \wedge \psi)$; i.e., $M \models \psi \to' \psi$.

The converse follows equally easily. Suppose $M \models \psi \to' \varphi$. Clearly if $M \models K\neg\psi$ then we trivially have that for all $u \in [\![\neg\varphi \wedge \psi]\!]_M$, there exists a world $v \in [\![\varphi \wedge \psi]\!]_M$ such that $v \succ u$ and $v$ dominates $[\![\neg\varphi \wedge \psi]\!]_M$. On the other hand, if $M \models \varphi \wedge \psi \gg \neg\varphi \wedge \psi$, then it follows by definition that for all $u \in [\![\varphi \wedge \psi]\!]_M$, there exists a world $v \in [\![\varphi \wedge \psi]\!]_M$ such that $v \succ u$ and $v$ dominates $[\![\neg\varphi \wedge \psi]\!]_M$. Either way we have $M \models \psi \to \varphi$. ∎

# References

Bendová, K., & Hájek, P. (1993). Possibilistic logic as tense logic. In Carreté, N. P., & Singh, M. G. (Eds.), *Qualitative Reasoning and Decision Technologies*, pp. 441–450.

Boutilier, C. (1994). Conditional logics of normality: a modal approach. *Artificial Intelligence, 68*, 87–154.

Brass, S. (1991). Deduction with super-normal defaults. In *Proceedings of 2nd International Workshop on Nonmonotonic and Inductive Logics*, Lecture Notes in AI, Vol. 659, pp. 153–174 Berlin/New York. Springer-Verlag.

Burgess, J. (1981). Quick completeness proofs for some logics of conditionals. *Notre Dame Journal of Formal Logic, 22*, 76–84.

Cayrol, C., Royer, R., & Saurel, C. (1992). Management of preferences in assumption-based reasoning. In *IPMU '92 (4th International Conference on Information Processing and Management of Uncertainty in Knowledge-Based Systems)*, Lecture Notes in Computer Science, Vol. 682, pp. 13–22 Berlin/New York. Springer-Verlag.

Delgrande, J. P. (1994). A preference-based approach to default reasoning. In *Proceedings, Twelfth National Conference on Artificial Intelligence (AAAI '94)*, pp. 902–908.

Dershowitz, N., & Manna, Z. (1979). Proving termination with multiset orderings. *Communications of the ACM, 22*(8), 456–476.

Doyle, J., Shoham, Y., & Wellman, M. P. (1991). A logic of relative desire. In *Proc. 6th International Symposium on Methodologies for Intelligent Systems*, pp. 16–31.

Dubois, D., & Prade, H. (1990). An introduction to possibilistic and fuzzy logics. In Shafer, G., & Pearl, J. (Eds.), *Readings in Uncertain Reasoning*, pp. 742–761. Morgan Kaufmann, San Francisco, Calif.

Dubois, D., & Prade, H. (1991). Possibilistic logic, preferential models, non-monotonicity and related issues. In *Proc. Twelfth International Joint Conference on Artificial Intelligence (IJCAI '91)*, pp. 419–424.

Fariñas del Cerro, L., & Herzig, A. (1991). A modal analysis of possibilistic logic. In *Symbolic and Quantitative Approaches to Uncertainty*, Lecture Notes in Computer Science, Vol. 548, pp. 58–62. Springer-Verlag, Berlin/New York.

Fine, T. L. (1973). *Theories of Probability*. Academic Press, New York.

Freund, M. (1993). Injective models and disjunctive relations. *Journal of Logic and Computation*, *3*(3), 231–247.

Friedman, N., & Halpern, J. Y. (1994). On the complexity of conditional logics. In *Principles of Knowledge Representation and Reasoning: Proc. Fourth International Conference (KR '94)*, pp. 202–213. Morgan Kaufmann, San Francisco, Calif.

Friedman, N., & Halpern, J. Y. (1997). Plausibility measures and default reasoning. *Journal of the ACM*. Accepted for publication. A preliminary version of this work appeared in *Proc. National Conference on Artificial Intelligence (AAAI '96)*, 1996, pages 1297–1304.

Gärdenfors, P. (1975). Qualitative probability as an intensional logic. *Journal of Philosophical Logic*, *4*, 171–185.

Geffner, H. (1992). High probabilities, model preference and default arguments. *Mind and Machines*, *2*, 51–70.

Goldszmidt, M., & Pearl, J. (1992). Rank-based systems: A simple approach to belief revision, belief update and reasoning about evidence and actions. In *Principles of Knowledge Representation and Reasoning: Proc. Third International Conference (KR '92)*, pp. 661–672. Morgan Kaufmann, San Francisco, Calif.

Halpern, J. Y., & Rabin, M. O. (1987). A logic to reason about likelihood. *Artificial Intelligence*, *32*(3), 379–405.

Hughes, G. E., & Cresswell, M. J. (1968). *An Introduction to Modal Logic*. Methuen, London.

Kraus, S., Lehmann, D., & Magidor, M. (1990). Nonmonotonic reasoning, preferential models and cumulative logics. *Artificial Intelligence*, *44*, 167–207.

Lewis, D. K. (1973). *Counterfactuals*. Harvard University Press, Cambridge, Mass.

Przymusinski, T. (1987). On the declarative semantics of stratified deductive databses and logic programs. In Minker, J. (Ed.), *Foundations of Deductive Databases and Logic Programming*, pp. 193–216. Morgan Kaufmann, San Francisco, Calif.

Shafer, G. (1976). *A Mathematical Theory of Evidence*. Princeton University Press, Princeton, N.J.

Spohn, W. (1988). Ordinal conditional functions: a dynamic theory of epistemic states. In Harper, W., & Skyrms, B. (Eds.), *Causation in Decision, Belief Change, and Statistics*, Vol. 2, pp. 105–134. Reidel, Dordrecht, Netherlands.